CALM: A Causal Analysis Language Model for Tabular Data in Complex Systems with Local Scores, Conditional Independence Tests, and Relation Attributes

Zhenjiang Fan, Department of Adult Neurosurgery, School of Medicine, Stanford University, Stanford, California, 94305, United States.
Zengyi Qin, Department of Electrical Engineering and Computer Science, School of Engineering, Massachusetts Institute of Technology, Cambridge, Massachusetts, 02139, United States.
Yuanning Zheng, Department of Biomedical Data Science, School of Medicine, Stanford University, Stanford, California, 94305, United States.
Bo Xiong, Department of Biomedical Data Science, School of Medicine, Stanford University, Stanford, California, 94305, United States.
Summer Han, Department of Adult Neurosurgery, School of Medicine, Stanford University, Stanford, California, 94305, United States.

# Abstract

Causal inference and discovery from observational data are fundamental to scientific fields like biology, where controlled experiments are often impractical. However, existing methods, including constraint-based (e.g., PC, causalMGM) and score-based approaches (e.g., NOTEARS), face significant limitations. These include an inability to resolve causal direction, restrictions to linear associations, sensitivity to violations of the faithfulness assumption, and inefficiency in searching vast hypothesis spaces. While large language models (LLMs) offer powerful reasoning capabilities, their application is hindered by a fundamental discrepancy: they are designed for text, while most causal data is tabular. To address these challenges, we introduce CALM, a novel causal analysis language model specifically designed for tabular data in complex systems. CALM leverages a Mamba-based architecture to classify causal patterns from pairwise variable relationships. It integrates a comprehensive suite of evidence, including local causal scores, conditional independence tests, and relational attributes, to capture a wide spectrum of linear, nonlinear, and conditional causal mechanisms. Trained on a diverse corpus of synthetic data (from linear, mixed, and nonlinear models) and 10 real-world biological datasets with rigorously validated causal relationships, our model ensures robustness and generalizability. Empirical evaluation demonstrates that CALM significantly outperforms existing methods in both simulation studies, achieving over 91% accuracy, and in a real-world application identifying causal factors in Hepatitis C virus progression. This work represents a significant step towards accurate and generalizable causal discovery by successfully adapting the pattern recognition capabilities of language models to the intricacies of tabular data.

# Introduction

Causal discovery from observational data is a cornerstone of scientific inquiry, particularly in fields like biology where controlled experiments are often infeasible. While numerous methods have been developed, they frequently face significant limitations that constrain their application to complex, real-world datasets.

Constraint-based algorithms, such as PC (Spirtes, Peter & Glymour, 1991) and causalMGM (Ge et al., 2020) are widely used but often produce graphs with abundant bidirectional edges, failing to resolve causal direction. Furthermore, methods like causalMGM are typically restricted to linear associations, while many biological processes are inherently nonlinear. Score-based approaches, including DL-based methods like DAG-GNN (Yu et al., 2019) and NOTEARS (Zheng et al., 2018), search over the vast space of directed acyclic graphs (DAGs). This exhaustive search can be inefficient and prone to error, as true causal signals may be neutralized by spurious ones within the large hypothesis space. A common vulnerability across many constraint-based methods is their reliance on the faithfulness assumption (Spirtes, Peter et al., 2000). This assumption, which posits that all conditional independencies in the data are due to the causal graph's structure, is often violated in practice by non-causal sources such as unmeasured confounding or selection bias, leading to the incorrect removal of edges. Collectively, these limitations mean that existing algorithms often excel only in specific niches, performing well on linear relationships or continuous variables, for instance, while struggling with the data heterogeneity, nonlinear interactions, and high dimensionality characteristic of biological systems.

Meanwhile, large language models (LLMs) have demonstrated remarkable reasoning and pattern recognition capabilities, offering a promising new paradigm for causal inference. However, a significant challenge remains: while most causal estimation data is tabular, LLMs are predominantly designed for and trained on textual data. This fundamental discrepancy hinders their direct application and effectiveness in discerning causal structures from datasets common in scientific research.

To address these challenges, we propose CALM, a novel causal analysis language model specifically designed for tabular data in complex biological systems. Building upon our previous work, CALM is architected to overcome the limitations of existing approaches by: 1) harnessing the powerful classification abilities of language models to identify nuanced causal patterns in pairwise variable relationships; 2) combining local causal scores, conditional independence tests, and association attributes to capture a wide spectrum of linear, nonlinear, and conditional causal mechanisms; 3) incorporating a diverse range of synthetic and real-world biological datasets during training to enhance robustness and generalizability beyond narrow data types; and 4) designing a modular and extensible system that allows users to integrate proprietary data and incorporate additional causal scores or statistical tests.

The following sections elaborate on our proposed causal estimation language model and present its empirical evaluation. The Methods section details the model's architecture and training procedure, while the Results section demonstrates its superior performance against existing baselines across both simulated and real-world datasets.

# Methods

This work aims to create a causal estimation language model for tabular data that can 1) leverage classification capabilities of language models to identify causal patterns in pairwise relationships; 2) integrate local causality estimation scores, conditional independence tests, and relation attributes, enabling it to capture a wide range of causal mechanisms; 3) incorporates a diverse range of simulation data and real-world biological datasets into the training procedure, ensuring robustness and generalizability; be also designed to be extensible, allowing users to integrate their own data and additional scores or tests. The output of our model is a DAG G=(V,E), where V is the set of nodes (variables) and E is the set of directed edges (causal relationships). Figure 1 provides an overview of the framework based on the proposed causal estimation language model. The framework comprises three main components: (a) the collection of diverse relation attributes, local causal scores, and conditional independence tests (top left), (b) the model training procedure (top right), and (c) the causal estimation process applied to a new data matrix using the trained model (bottom). The following subsections detail each of these components.

## 2.1 Causal Estimation Language Model

### Language Model

The language model architecture utilized in this study is Mamba, proposed by Gu and Gao et al. (Gu & Dao, 2023). The Mamba architecture is primarily based on a selective structured state space model (SSM). The key innovation is that the parameters of this SSM are dynamically computed based on the input, allowing the model to perform content-based reasoning, meaning it can selectively choose to remember or forget information depending on the current token. Figure 1 provides an overview of the proposed Causal Estimation Language Model (**CausalML**). The model input is a score data matrix, $X \in R^{m \times k}$, retrieved from the score collection process (bottom of Figure 1). Here, $m$ denotes the number of pairwise relationships and $k$ represents the number of local scores, tests, and relational attributes.

Following the tabular data processing approach proposed by Thielmann et al. (Thielmann et al., 2024), the input data is first encoded into an embedded representation. Categorical features are encoded using distinct, feature-specific vocabularies to prevent issues arising from binary or integer encoding. An `<UNK>` token is included in each vocabulary to handle unknown or missing values during both training and inference. For numerical features, we apply Periodic Linear Encodings (Gorishniy et al., 2021). The bin boundaries for these encodings are determined using simple decision trees, and each feature is subsequently passed through a linear layer for rescaling. The resulting embedded representations are denoted as $Z \in R^{N \times J \times d}$ ($J$ is the number of features and $d$ is the embedding dimension)to distinguish them from the raw input features $X$. Finally, the combined embeddings $Z$ are divided into batches before they are processed by a stack of Mamba layers.

As illustrated in Figure 1, the first layer of these Mamba blocks is an one-dimensional convolutional layer to account for invariance of feature ordering in the pseudo-sequence. The output of the convolutional layer will have the same shape as the input if padding is set to the kernel size -1.

This output will be sent to a state-space model (SSM). In the SSM, each hidden state $h_j \in R^{N \times d \times \delta}$ is iteratively updated by the following formula:

$$h_j = exp(\Delta \odot_3 A)_{:,j,:,:} \odot_{1,2,3} h_{j-1} + ((\Delta \odot_{1,2} B) \odot_{1,2} \bar{z})_{:,j,:,:}$$

where $\delta$ is an inner dimension; $\bar{z} \in R^{N \times J \times d \times 1}$ has a shape of N×J×d×; $\Delta \in R^{N \times J \times d \times 1}$ is the gating matrix which modulates the contributions of the state transition and the current input, controlling the extent to which the previous hidden state is updated; $\odot$ denotes the element-wise product with broadcasting over any singleton dimensions; $A \in R^{1 \times 1 \times d \times \delta}$ is the state transition matrix that governs the evolution of the hidden state from one time step to the next; $B \in R^{N \times J \times 1 \times \delta}$ projects the input features into the hidden state space, governing their influence on the state update at each time step.

Then, the output representation $\hat{x}$ is calculated using the following formula:

$$\hat{x} = (H \cdot_4 C) + (\alpha \odot_3 z)$$

where $H$ is retrieved by stacking all the hidden states ($H = [h_0, h_1, ..., h_{T-1}] \in R^{N \times J \times d \times \delta}$); $C \in R^{N \times J \times 1 \times \delta}$ and $\alpha \in R^{1 \times 1 \times d}$ are learnable parameters.

Then, $\hat{x}$ is passed to the final linear layer. The output of this linear layer is computed as follows:

$$\hat{x}_{final} = (\hat{x} \odot_{1,2,3} z') W_{final} + b_{final}$$

After applying pooling along the feature axis, the output of the final linear layer $\hat{x}_{final}$ is sent to a task-specific model head (Thielmann et al., 2024), in this case, a classification task. Therefore, the loss is calculated using the binary cross entropy.

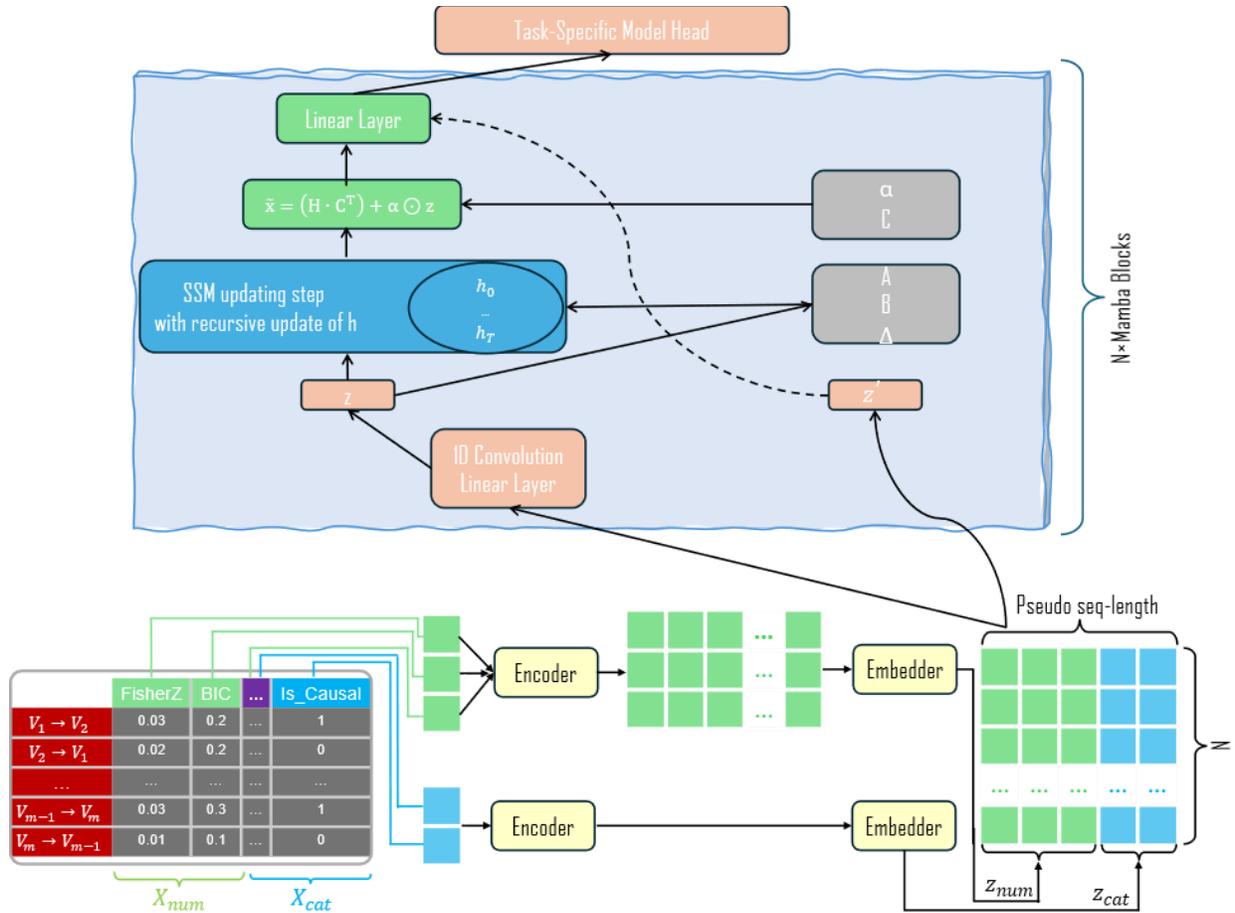

*Figure 1. A schematic overview of the Causal Analysis Language Model (CALM). The pipeline begins with the encoding and embedding of the input score data matrix (bottom), followed by its processing through the subsequent language model architecture (top).*

## Score Collecting

Real-world observational datasets are characterized by diverse data types and complex relationships, necessitating analytical models that can integrate multiple sources of evidence. Our approach, illustrated in the top left of Figure 1, addresses this by aggregating a wide spectrum of local causal scores, conditional independence tests, and relational attributes. This comprehensive collection enables the capture of diverse causal mechanisms for any pairwise relationship.

As illustrated in Figure 2, the process takes as input a data matrix $X_{orig} \in R^{m \times k}$ (where $m$ is the number of observations and $k$ is the number of variables) and a set of pairwise relationships $S$. The specification of $S$ defines the task:

**Causal Inference**: If $S$ consists of user-defined pairs (e.g., all relationships with a target variable $R$), the goal is to infer causal directions for these specific relation pairs.

**Causal Discovery**: If $S$ is not predefined, the task is to learn the complete underlying causal graph from $X_{orig}$.

Prior to the process, $X_{orig}$ is normalized to ensure all collected scores are on a consistent scale. Pairs that satisfy conditional independence (i.e., show no evidence of a causal relationship) are classified as non-causal and excluded from further scoring.

The collected information is organized into three complementary categories, providing a multi-faceted view of each pairwise relationship. A complete list of all collected metrics is provided in **Supplemental Table 2**.

1. Conditional Independence Tests: These tests are fundamental for distinguishing causal associations from spurious ones by testing if two variables are independent given a conditioning set. Our compiled suite includes the Fisher-Z test (FISHER, R. A., 1921), Hilbert-Schmidt Independence Criterion (HSIC) (Gretton et al., 2005), Kernel-based Conditional Independence test (KCI) (Zhang et al., 2012), and their variants.

2. Contextual Attributes: The performance of causal discovery methods is often dependent on data characteristics. To inform the model of this context, we collect attributes such as variable data types (continuous/discrete) and pairwise relationship types (linear/nonlinear), enabling it to weight evidence appropriately across diverse data scenarios.

3. Causal Direction Estimators: This category comprises scores that directly inform the likely direction of a causal relationship. We integrate a diverse set of estimators, including the Additive Noise Model (ANM) (Hoyer et al., 2008), Bayesian Information Criterion (BIC) (Schwarz, 1978), scores based on information theory (Mutual Information, Conditional Mutual Information) (Kozachenko, 1987; Kraskov et al., 2004; Ross, 2014), and model fit metrics (e.g., negative log-likelihood) (Huang, 2018).

By integrating evidence from these three categories, our causal estimation language model is equipped to uncover the complex patterns underlying data generation processes.

The output of the score-collecting process is a data matrix $X \in R^{n \times d}$ (where $n$ is the number of pairwise relationships and $d$ is the number of local scores, tests, and relation attributes).

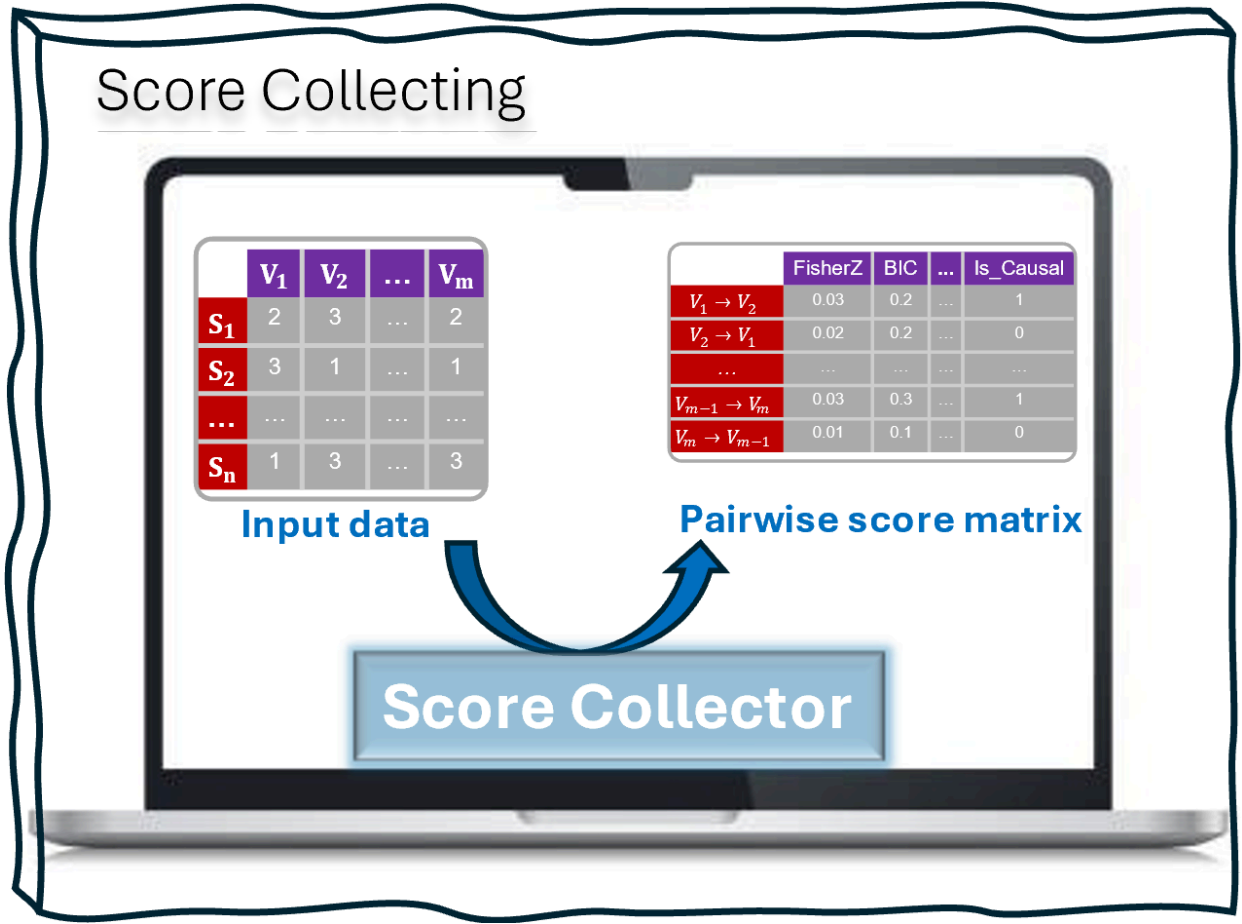

*Figure 2. Schematic of the score collection pipeline, transforming an input feature matrix (samples × features) into an output matrix of pairwise relationship scores (relationships × scores/tests/attributes).*

## Causal Data Simulation

To enhance the robustness and generalizability of CALM, we integrate diverse synthetic data from three primary sources: a linear Gaussian model, a mixed graphical model, and a suite of nonlinear functions.

1. Linear Gaussian (Fisher) Model
We simulate linear Gaussian data based on the foundational work of Fisher (Fisher, Ronald A., 1936), implemented using the Tetrad software suite [2]. This model generates datasets with linear relationships and Gaussian noise. Each dataset comprises 50 variables and 1,500 samples, with all parameters set to Tetrad's default values.

2. Mixed Graphical Model (MGM)
To simulate data with both continuous and discrete variables, we employ the Mixed Graphical Model (MGM) proposed by Lee & Hastie (Lee & Hastie, 2015), also implemented in Tetrad (Scheines et al., 1998). The MGM captures three types of pairwise causal relationships:
Continuous-Continuous: Modeled with a multivariate Gaussian distribution.
Discrete-Discrete: Modeled with a discrete pairwise Markov Random Field (MRF).

Mixed: Relationships between a continuous and a discrete variable.
Each generated dataset contains 50 variables and 1,500 samples, using Tetrad's default parameters.
3. Nonlinear Functions:
Nonlinear relationships are critical for modeling complex systems. We generate data using a suite of functions, including arctangent, sine, cosine, hyperbolic tangent, and power functions, to emulate a wide spectrum of behaviors such as periodic oscillations, sigmoidal transitions, and power-law dynamics. This provides a robust testbed for evaluating model performance under controlled nonlinearities. Together, these simulations create realistic and complex datasets that mimic real-world phenomena.

## Real-World Datasets and Causal Relationship Validation

To enhance the robustness and generalizability of CALM, we augment our synthetic data (from Linear Fisher, Mixed Graphical, and nonlinear models) with a diverse collection of real-world biological datasets. The efficacy of language models is contingent on the quality, diversity, and scale of training data. Real-world data is paramount for ensuring models generalize to unseen scenarios, as it encapsulates the inherent noise, variability, and complexity of practical applications.

We curated 10 prominent real-world datasets (detailed in **Supplemental Table 1**), several of which are well-established benchmarks for evaluating deep learning models. This collection includes six clinical/laboratory datasets, one long-read sequencing dataset, and three single-cell sequencing datasets. For clinical datasets with missing values, we applied imputation strategies including mean imputation and K-Nearest Neighbors.

To ensure high-confidence causal relationships for training, we imposed a rigorous three-step filtration criteria. Each relationship was required to:
1. Demonstrate Statistical Association: Pass a Fisher-z conditional independence test (FISHER, R. A., 1921).
2. Be Identified as Causal by Established Methods: Be consistently identified by three distinct causal discovery algorithms: the PC algorithm (Spirtes, Peter et al., 2000), the Fast Causal Inference (FCI) (Spirtes, Peter L. et al., 2013), and the Greedy Equivalence Search (GES) (Chickering, 2002).
3. Have Literature Support: Be corroborated by evidence in published scientific literature.

For relationships satisfying the first two criteria, we performed a systematic literature review using Scite AI's tool (for reproducibility) and Google Scholar to identify supporting publications.

## Model Training

To enhance the robustness and generalizability of the final trained model, we integrate a diverse range of simulation data alongside 10 real-world datasets during the training process (as illustrated in **Figure 3**). The following steps outline our approach to obtaining the final trained model.

**Simulation Data**: We generated five simulation datasets (see **Causal Data Simulation**), each comprising 1,500 observations, 50 variables, and 100 predefined causal relationships. These datasets include one generated using the Linear Fisher Model (Fisher, Ronald A., 1936), three using the Mixed Graphical Model (Lee & Hastie, 2015), and one nonlinear dataset. Additionally, we incorporated 10 real-world datasets with validated causal relationships (see **Real-World Datasets and Causal Relationship Validation**). All datasets were rescaled using min-max normalization.

**Collecting Score Data**: For each dataset, we collected local scores, conditional independence test results, and attributes for every pairwise relationship (see **Score Collecting**). To create a balanced training set, we introduced an equivalent number of false causal relationships into the score data, derived from the p-values of conditional independence tests. These false relationships served as negative controls for model training. Subsequently, we merged the score data across all datasets to create a unified training dataset.

To enhance model robustness and generalizability, we integrated a diverse corpus of data for training, comprising both simulated and real-world datasets (**Figure 3**).

The training data consists of two components:
- **Simulated Data**: We generated five distinct datasets, each containing 1,500 observations over 50 variables with 100 predefined causal relationships (see **Causal Data Simulation**). These include one linear dataset (generated using the Linear Fisher Model (Fisher, Ronald A., 1936)), three datasets from a Mixed Graphical Model (Lee & Hastie, 2015), and one nonlinear dataset.
- **Real-World Data**: We incorporated 10 real-world datasets with validated causal structures (see **Real-World Datasets and Causal Relationship Validation**).

All datasets were normalized using min-max scaling prior to collecting score data. For every pairwise relationship within each dataset, we computed a feature vector comprising local scores, conditional independence tests, and relational attributes (see **Score Collecting**). To create a balanced training set, we introduced an equal number of non-causal (false) relationships as negative controls. These were systematically generated from the p-values of conditional independence tests. As illustrated in **Figure 3**, the score data from all simulated and real-world datasets were subsequently merged into a unified training dataset. The key hyperparameters of the model architecture are summarized in **Table 1**. To ensure full reproducibility, the final trained model, including its architecture, learned parameters, and optimizer state, is saved in a standardized format. The complete source code and detailed documentation are publicly available in our GitHub repository: https://github.com/ZhenjiangFan/CALM.

*Table 1: Model Architecture Hyperparameters*

| Hyperparameter | Value |
| --- | --- |
| Dimensionality (Hidden Size) | 64 |
| Number of Layers | 4 |
| Feed-Forward Expansion Factor | 2 |
| Convolution Kernel Size | 4 |
| State Dimensionality (Recurrent) | 128 |
| Pooling Method | Average |

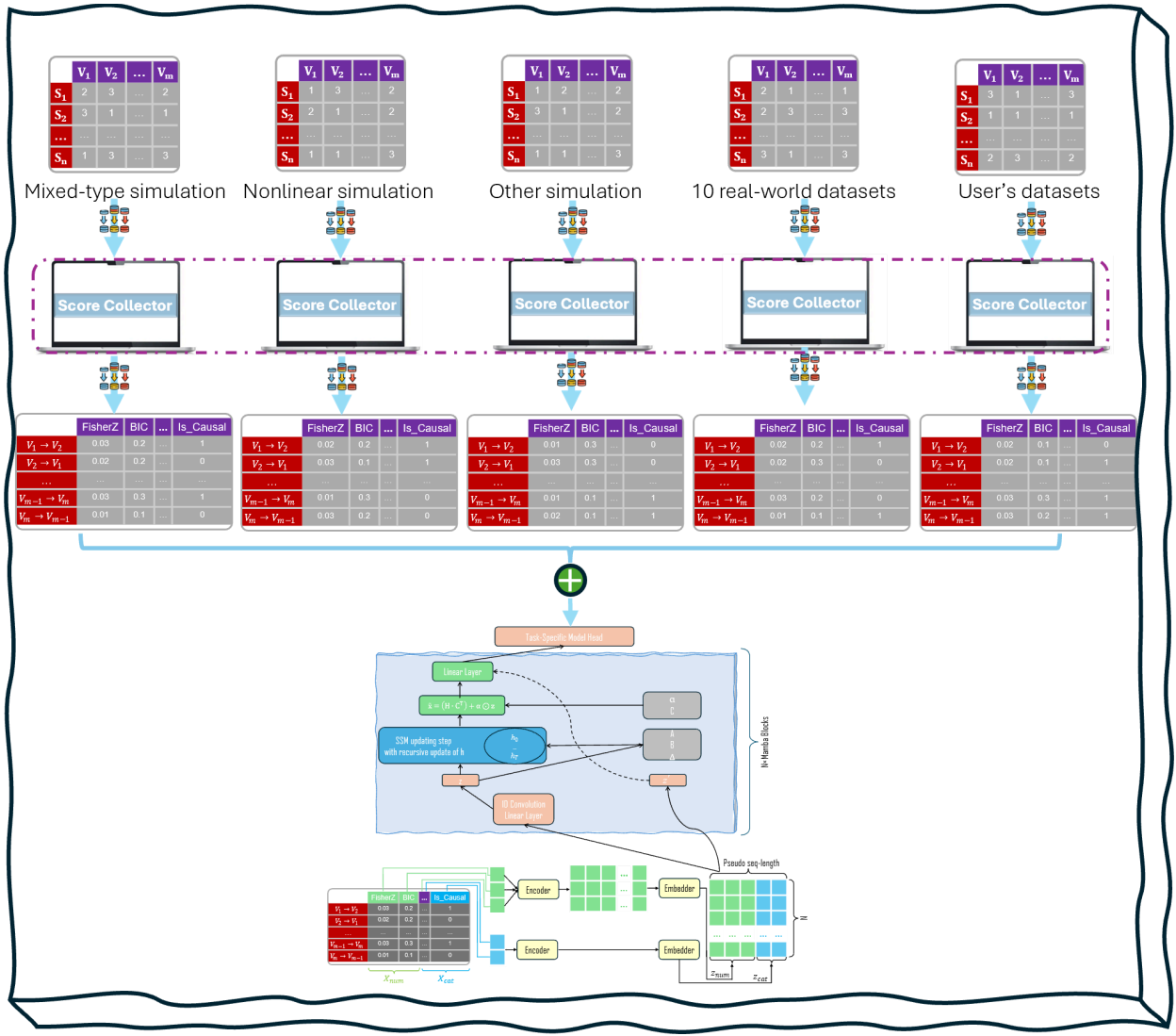

# Causality Estimating

The causal estimation process, illustrated in **Figure 4**, operates on an input data matrix $X \in R^{m \times k}$ (m observations and k variables) and a set of pairwise relationships $S$ to be evaluated. The set $S$ is defined according to the following task-specific scenarios:

**Targeted Causal Inference**: $S$ contains all relationships associated with a user-specified target variable.

**Focused Causal Inference**: $S$ is a user-defined set of specific relationships of interest.

**Causal Discovery**: $S$ comprises all possible pairwise relationships among the $k$ variables.

The first two scenarios represent causal inference tasks, whereas the third constitutes a full causal discovery procedure. The subsequent workflow consists of six sequential steps designed to output a DAG.

**Step 1: Data Normalization**. The input matrix X is normalized using min-max scaling to ensure all variables share a consistent scale.

**Step 2: Score Data Collection**. For every pairwise relationship in S, we compute a feature vector comprising local scores, conditional independence tests, and relational attributes. Relationships that fail to meet a predefined conditional independence threshold are filtered out, retaining only statistically plausible candidates for model inference.

**Step 3: Model Inference**. The filtered score data is passed as input to the pre-trained causal language model. The model estimates the likelihood of a causal relationship for each pair; directed edges are added to an initial graph for all relationships classified as causal.

**Step 4: Graph Construction**. A directed graph is formally constructed from the edges identified in Step 3.

**Step 5: Orienting Bidirectional Edges**. Potential bidirectional edges are resolved by evaluating the two possible causal directions. The direction assigned a higher confidence score by the model is selected, and the alternative is discarded.

**Step 6: Cycle Removal and DAG Output**. The directed graph is processed to detect and break any cycles, ensuring acyclicity. This is achieved by iteratively removing the edge with the lowest causal confidence score within each identified cycle. The final output is a valid DAG representing the estimated causal structure.

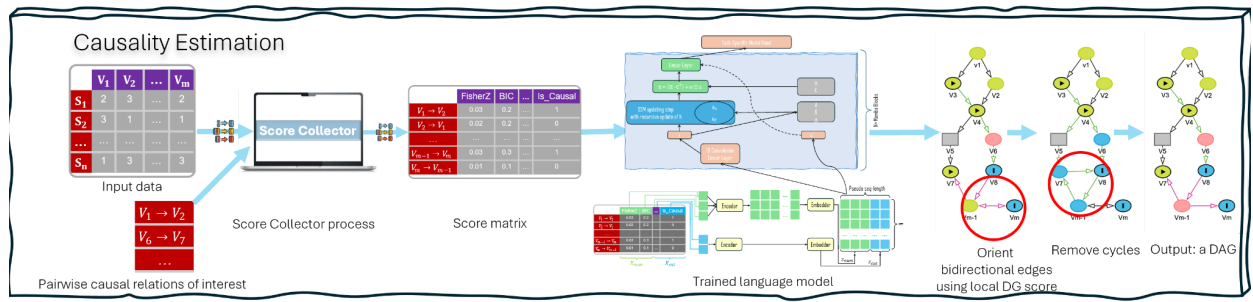

*Figure 4. The CALM workflow for causal structure estimation. The pipeline consists of six sequential steps: (1) input data normalization, (2) collection of scores and conditional independence tests, (3) causal relationship inference via the trained model, (4) construction of an initial directed graph, (5) resolution of bidirectional edges, and (6) cycle removal to produce a final DAG.*

# Results

## 3.1 Simulation Results

To evaluate the performance of CALM, we conducted extensive simulations and compared it against three established baseline methods: GES, FCI, and PC. The evaluation focused on prediction accuracy and precision.

Our simulation study employed 20 datasets, each comprising 1,500 samples over 90 variables (see **Causal Data Simulation**). Each dataset was constructed with a ground-truth causal graph containing 180 edges: 60 linear, 60 nonlinear, and 60 mixed-type causal relationships. The baseline methods were executed using implementations from the widely-used Tetrad causality package (https://github.com/cmu-phil/tetrad).

As illustrated in **Figure 5**, CALM demonstrated consistently superior performance across all datasets, achieving a mean accuracy rate above 91%. In contrast, the mean accuracy rates for GES, FCI, and PC were approximately 63%, 50%, and 69%, respectively. All methods exhibited stable performance with minimal variance across the simulated datasets. We note that the PC algorithm, utilizing the latest implementation from Tetrad, outperformed both GES and FCI.

In summary, our simulation results demonstrate that CALM significantly outperforms state-of-the-art causal discovery methods in accuracy under complex, noisy conditions. This robust performance underscores the effectiveness of our approach and its potential for practical applications in causal inference and discovery.

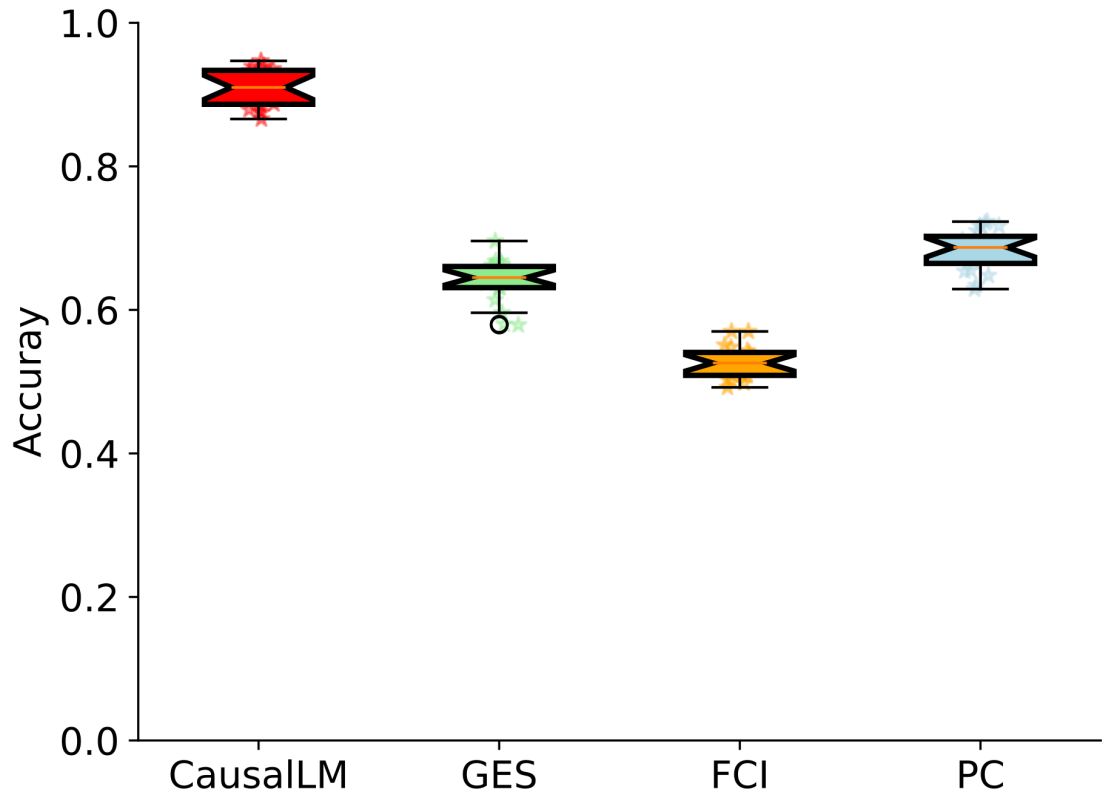

*Figure 5. Performance comparison between DeepCEF and existing methods (GES, FCI, and PC) in terms of prediction accuracy and precision. DeepCEF outperforms GES, FCI, and PC in predicting the number of true causal relationships across 20 simulation datasets.*

## 3.2 Real-world Application Results

We assessed the performance of CALM using a real-world Hepatitis C Virus (HCV) dataset (n=615, 12 features) (Hoffmann et al., 2018). The analysis aimed to identify causal factors to the progression of hepatitis C. The target variable encompasses a spectrum of conditions, ranging from a negative class (Blood Donor/Suspect Blood Donor) to a series of stages marking disease progression: Chronic Hepatitis C, Liver Fibrosis, and Cirrhosis (Anderson et al., 2000; Langohr et al., 2008).

**Figure 6** illustrates the causal graph identified by CALM for the hepatitis C progression variable. The model identified key biomarkers associated with liver function, such as Alanine Aminotransferase (ALT) and Aspartate Transaminase (AST) (Akkaya et al., 2007; Amjad et al., 2021; Giannini et al., 2003). Elevated levels of these enzymes in the bloodstream are clinical indicators of liver damage. While not direct causes of HCV infection, they are implicated in the disease's progression, which is the focus of the target variable. Furthermore, patient age was identified as a factor, consistent with clinical knowledge that older adults often experience accelerated disease progression (Naggie, 2017). Other markers, including albumin and bilirubin, were also found to be causally related to the progression stages (Ewid et al., 2025; Fujita et al., 2019; Martínez Herreros et al., 2022). In conclusion, the causal graph suggests that the progression of HCV is driven not by a single cause, but by a network of interacting factors related to liver function and patient demographics.

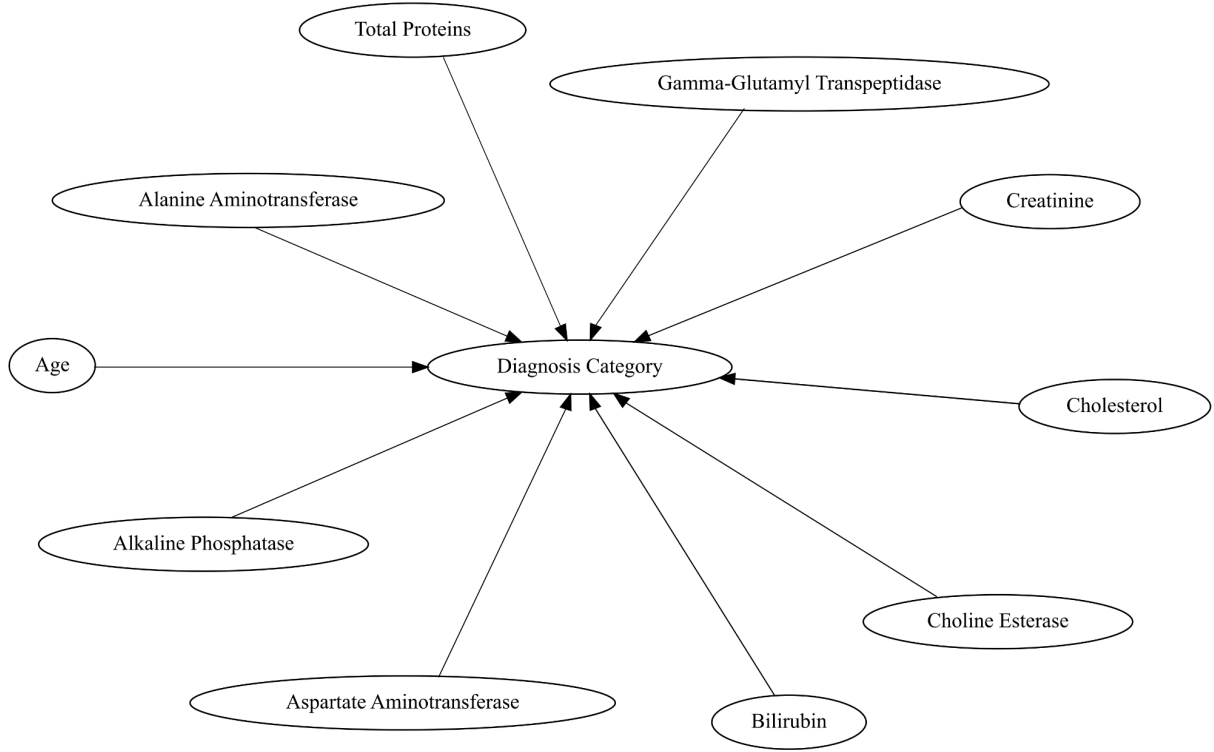

*Figure 6. Causal relationships identified by CALM for the Hepatitis C Virus (HCV) category variable.*

# Conclusion

In this study, we introduced CALM, a novel causal analysis language model specifically designed to overcome the limitations of existing methods when applied to complex, real-world data. CALM addresses key challenges in causal inference and discovery by leveraging the pattern recognition capabilities of language models on tabular data, integrating diverse causal signals (local scores, independence tests, and association attributes), and ensuring robustness through training on a wide range of synthetic and real-world datasets. Furthermore, its modular architecture provides a flexible and extensible framework for the scientific community.
Our empirical evaluation demonstrated that CALM consistently outperforms existing baselines on both simulated benchmarks and real-world biological datasets. These results confirm that the model effectively captures a broader spectrum of causal relationships, including linear, nonlinear, and conditional mechanisms, than methods constrained by specific data type assumptions or reliance on the faithfulness condition.

The development of CALM represents a significant step towards more accurate and generalizable causal discovery in heterogeneous systems. By successfully adapting language models to the intricacies of tabular data, this work opens new avenues for leveraging advanced AI architectures in scientific domains where understanding causality is paramount.

# Acknowledgments

The authors acknowledge the computational support provided by Google Cloud Research Credits from GCP Education Programs and AMD's High Performance Compute Fund.

## Supplemental Tables

*Supplemental Table 1. The table provides details of the 10 real-world biological datasets utilized in the training process, including disease type, data type, data source, access links, and other relevant information.*

| Disease | Description | Source |
| --- | --- | --- |
| Chronic kidney disease | Clinical and laboratory data | (Rubini et al., 2015) |
| Diabetes | Clinical and laboratory data | (National Institutes of Health, 2010) |
| Ovarian cancer | Clinical and laboratory data | (Lu et al., 2020) |
| Sepsis | Clinical and laboratory data | (Reyna et al., 2020) |
| Health and nutrition survey | Clinical and laboratory data | cdc.gov |
| Smoking status | Clinical and laboratory data | data.go.kr |
| Tumors | Long-read sequencing data | (O'Neill et al., 2024) |
| Breast cancer | Single-cell sequencing data | (Asemota et al., 2024) |
| Lung cancer | Single-cell sequencing data | Yang, Xu, Zhou, Liu et al. |

| Diabetes | Single-cell sequencing data | (Qian et al., 2024) |
|----------|----------------------------|---------------------|

*Supplemental Table 2. Overview of the three categories of information collected in this study: (1) (conditional) independence tests, (2) context information, and (3) causal direction estimators. The table details their applications in causal inference, associated authors or references, and their potential value ranges.*

| Name | Usage | Source |
|------|-------|--------|
| Fisher-z | Conditional independence test | (Fisher, Ronald A., 1936) |
| Fisher-z (binary) | Binary label for Fisher-z conditional independence test | (Fisher, Ronald A., 1936) |
| HSIC statistic | Independence test | (Gretton et al., 2005) |
| HSIC p-value | HSIC Independence test | (Gretton et al., 2005) |
| HSIC (binary) | Binary label for HSIC Independence test | (Gretton et al., 2005) |
| KCI test | Conditional independence test | (Zhang et al., 2012) |
| KCI (binary) | Binary label for KCI test | (Zhang et al., 2012) |
| Missing-value Fisher-z | Conditional independence test | (Fisher, Ronald A., 1936) |
| Missing-value Fisher-z (binary) | Binary label for Missing-value Fisher-z test | (Fisher, Ronald A., 1936) |
| Negative k-fold cross-validated log likelihood (with regularization parameter of 0.01) | Local score | (Huang, 2018) |
| Negative k-fold cross-validated log likelihood (with regularization parameter of 0.05) | Local score | (Huang, 2018) |
| Negative k-fold cross-validated log likelihood (with regularization parameter of 0.1) | Local score | (Huang, 2018) |
| Negative k-fold cross-validated log likelihood (binary) | Local score | (Huang, 2018) |
| Cause variable data type (discrete or not) | Variable attribute | |
| Cause variable data type (discrete or not) | Variable attribute | |
| MI | Mutual information between cause and effect | (Kraskov et al., 2004) |
| CMI | Conditional mutual information between cause and effect | (Ver Steeg & Galstyan, 2015) |
| ANM | Estimating cause and effect between two variables | (Hoyer et al., 2008) |
| LM test | Estimating linearity between two variables | statsmodels.org |
| LM test (p-value) | Determining linearity between two variables | statsmodels.org |
| LM test (p-value, augmentation type 'fitted', F-test) | Determining linearity between two variables | statsmodels.org |
| LM test (p-value, augmentation type 'exog', F-test) | Determining linearity between two variables | statsmodels.org |
| LM test (p-value, augmentation type 'princomp', F-test) | Determining linearity between two variables | statsmodels.org |
| LM test (p-value, augmentation type 'fitted'', chi-square test) | Determining linearity between two variables | statsmodels.org |
| LM test (p-value, augmentation type 'exog'', chi-square test) | Determining linearity between two variables | statsmodels.org |
| LM test (p-value, augmentation type 'princomp'', chi-square test) | Determining linearity between two variables | statsmodels.org |
| BIC (lambda 0.1) | Determining causal direction for a relation | (Schwarz, 1978) |
| BIC (lambda 0.5) | Determining causal direction for a relation | (Schwarz, 1978) |
| BIC (lambda 1) | Determining causal direction for a relation | (Schwarz, 1978) |
| BIC (lambda 1.5) | Determining causal direction for a relation | (Schwarz, 1978) |
| BIC (binary, lambda 1) | Determining causal direction for a relation | (Schwarz, 1978) |
| DG | Determining causal direction for a relation | (Andrews et al., 2019) |
| DG (binary) | Determining causal direction for a relation | (Andrews et al., 2019) |